\documentclass[conference]{IEEEtran}

\makeatletter
\def\ps@headings{%
\def\@oddhead{\mbox{}\scriptsize\rightmark \hfil \thepage}%
\def\@evenhead{\scriptsize\thepage \hfil \leftmark\mbox{}}%
\def\@oddfoot{}%
\def\@evenfoot{}}
\makeatother
\usepackage{soul}
\usepackage{wrapfig}
\usepackage{mwe}
\usepackage{booktabs}
\usepackage{graphicx}
\usepackage{multirow}
\usepackage{subfigure}
\usepackage{booktabs}
\usepackage{xcolor}
\usepackage[colorlinks]{hyperref}
\usepackage{color, colortbl}
\definecolor{Gray}{gray}{0.9}
\usepackage{array,multirow}
\usepackage{float}
\usepackage[utf8]{inputenc}
\usepackage{dirtytalk}
\usepackage{cite}
\usepackage{array,multirow}
\usepackage{epsfig,bm,amsmath,amssymb,graphicx,setspace}
\usepackage{algorithm}
\usepackage{algpseudocode}
\usepackage{color,placeins}
\usepackage{multirow}

\usepackage{tabularray}

\usepackage{enumerate}
\usepackage{turnstile}
\usepackage{caption}
\usepackage{arydshln,paralist}
\usepackage{url}
\usepackage{booktabs}
\usepackage{array}
\usepackage{soul,tabularx, booktabs}
\usepackage{xcolor}
\usepackage{verbatim}
\usepackage[colorinlistoftodos]{todonotes} 
\usepackage{balance}

\usepackage{makecell}

\definecolor{usethiscolorhere}{rgb}{0.86666,0.78431,0.78431}

\makeatother
\begin{document}
\title{RAISE: Realness Assessment for Image Synthesis and Evaluation}

\author{
    Aniruddha Mukherjee\IEEEauthorrefmark{1},
    Spriha Dubey\IEEEauthorrefmark{2},
    Somdyuti Paul\IEEEauthorrefmark{3}\\[1ex]
    \IEEEauthorrefmark{1}Department of Computer Science and Engineering, Kalinga Institute of Industrial Technology, Bhubaneswar, India\\
    mukh.aniruddha@gmail.com\\
    \IEEEauthorrefmark{2}Department of Metallurgical and Materials Engineering, Indian Institute of Technology Kharagpur, Kharagpur, India\\
    sprihadubey2004@gmail.com\\
    \IEEEauthorrefmark{3}Department of Artificial Intelligence, Indian Institute of Technology Kharagpur, Kharagpur, India\\
    somdyuti@ai.iitkgp.ac.in\\[1ex]
    \IEEEauthorrefmark{1} \IEEEauthorrefmark{2}Equal contribution
}
    
\thanks{\textsuperscript{*}Equal Contribution.}    
\IEEEoverridecommandlockouts

\IEEEpubid{\begin{minipage}{\textwidth}\ \\[12pt] \\
\copyright 2025 IEEE. Personal use of this material is permitted. Permission from IEEE must be obtained for all other uses, in any current or future media, including reprinting/republishing this material for advertising or promotional purposes, creating new collective works, for resale or redistribution to servers or lists, or reuse of any copyrighted component of this work in other works.\\ 
\end{minipage}}

\maketitle

\begin{abstract} 

The rapid advancement of generative AI has enabled the creation of highly photorealistic visual content, offering practical substitutes for real images and videos in scenarios where acquiring real data is difficult or expensive. However, reliably substituting real visual content with AI-generated counterparts requires robust assessment of the perceived realness of AI-generated visual content, a challenging task due to its inherent subjective nature. To address this, we conducted a comprehensive human study evaluating the perceptual realness of both real and AI-generated images, resulting in a new dataset, containing images paired with subjective realness scores, introduced as RAISE in this paper. Further, we develop and train multiple models on RAISE to establish baselines for realness prediction. Our experimental results demonstrate that features derived from deep foundation vision models can effectively capture the subjective realness. RAISE thus provides a valuable resource for developing robust, objective models of perceptual realness assessment.

\end{abstract}
\IEEEpeerreviewmaketitle
\begin{IEEEkeywords}
AI generated images, Perceptual realness assessment, Psychovisual human study, Feature design, CNN, Transfer learning
\end{IEEEkeywords}









\section{Introduction}

The emergence of diffusion based generative models has revolutionized the creation of synthetic visual data, with unprecedented levels of photorealism being attained by models in this class such as Stable diffusion~\cite{latentdiff} and DALL-E 2 ~\cite{dalle2}. The ability of such generative models to process complex textual and visual prompts to produce photorealistic results distinguishes them from their predecessors, such as those based on generative adversarial learning or variational autoencoders. The widespread availability of such sophisticated models that can create visual data that are virtually indistinguishable from photographic depictions of real scenes raises concerns regarding privacy violations, intellectual property issues, security threats, and the spread of misinformation. To address such concerns, researchers have focused on developing robust solutions to distinguish AI generated images and videos from real ones. 

Despite such challenges, photorealistic synthetic data could  be invaluable in numerous scenarios, where acquiring or accessing real data is difficult, expensive or intrusive. While large scale image and video datasets are available, public datasets of similar scales are quite rare for domains like biomedical imaging, as acquiring such images are often expensive as well as intrusive, while distributing them is likely to be precluded by protocols that are designed to protect the privacy of individual test subjects. Likewise, high-resolution satellite images, underwater images and astronomical images require highly specialized imaging equipment, making them costly to acquire. Further, even images and videos of natural scenes such as movie footage may be restricted by copyrights, limiting their use for most purposes. Such limitations on the amount of available data often stymies the development of robust deep learning models. Generative AI, capable of producing realistic and domain-specific visual data, offers a promising solution to this problem by augmenting the available data with generated samples. However, this strategy relies on the ability to quantify how closely synthetic data resemble real images. While humans can often identify subtle visual artifacts produced by AI, human evaluation is subjective and unscalable. Thus, developing reliable models to quantify perceptual realness is crucial for the practical deployment of AI generated visual content—a direction that remains underexplored.
 
\IEEEpubidadjcol To address the challenge of quantifying the perceptual realness of AI generated visual content, we introduce RAISE, a dataset of AI generated images annotated with their subjective realness scores collected via a systematic human study. As humans are the primary consumers of visual media, this dataset provides a valuable resource for training and benchmarking models that estimate perceived realness. Such models can further guide the fine-tuning of generative AI models to reduce visual inconsistencies and produce more realistic images. In this paper, we describe the construction of then RAISE dataset, analyze it statistically, and use it to build baseline models for realness prediction\footnote{Our dataset and realness prediction models are available at \url{https://github.com/annimukherjee/RAISE.git}}.


\section{Related Works}





Many image and video quality assessment models have been developed to grade the perceptual quality of natural images such as~\cite{niqe, nima}. The development and benchmarking of these models have been significantly aided by large-scale subjective quality datasets like~\cite{ava, koniq}. However, benchmarking the perceptual characteristics of AI-generated images (AIGIs) has received relatively little attention, with interest in this area emerging only recently. One of the earliest efforts to benchmark the perceptual quality of AIGIs, introduced a dataset of 1080 AIGIs annotated with subjective quality ratings~\cite{zhang2023perceptual}. In~\cite{li2023agiqa}, both perceptual quality and alignment scores for AIGIs were collected from human subjects to construct the AIGIQA-3K dataset. A set of over 2000 AIGIs were rated based on their quality, authenticity as well as alignment with the text prompts to constitute the AIGCIQA dataset~\cite{wang2023aigciqa2023}. The AGIN dataset~\cite{Agin} includes AIGIs annotated for technical quality and rationality, which were modeled separately and then fused to estimate image naturalness. More recently, AIGCIQA-20K~\cite{li2024aigiqa} introduced a large-scale dataset of 20,000 AIGIs with subjective ratings reflecting both visual quality and prompt alignment.

A variety of objective metrics have been developed to evaluate different aspects of AI-generated image (AIGI) content. Semantic alignment between an image and its generating text prompt is commonly assessed using Contrastive Language–Image Pre-training (CLIP) embeddings~\cite{clip}. Inception Score (IS)~\cite{inceptionscore} measures the clarity and diversity of generated images, while Fréchet Inception Distance (FID)~\cite{fid} and Kernel Inception Distance (KID)~\cite{kid} evaluate the similarity between the feature distributions of real and generated images. However, these metrics are designed for set-level evaluation and cannot assess or rank individual images.

Recently, the NTIRE challenge~\cite{ntire} focused on the quality assessment of AI-generated images and videos using the AIGI-20K~\cite{li2024aigiqa} and T2VQA-DB~\cite{kou2024subjective} datasets, respectively. The top performing model for the image track of the challenge measured the correspondence between AIGIs and text prompts to achieve a Spearman's rank order correlation  coefficient (SROCC) of 0.9076 with the ground truth mean opinion scores (MOS)~\cite{peng2024aigc}. In the video track, the leading approach combined aesthetic, technical and prompt alignment aspects to attain a SROCC of 0.8322~\cite{aigcvqa}. A novel metric proposed in~\cite{glips} used transformer-based attention to measure local similarity and maximum mean discrepancy for global distributional alignment, showing strong correlation with human judgments of photorealism, quality, and prompt alignment.

Most existing image/video quality assessment (I/VQA) models for AI generated content are designed to score content based on its alignment with the associated text prompt. Consequently, these models cannot be applied when prompts are unavailable. Relatively fewer works address quantifying the realness of AIGIs as a standalone problem that does not rely on the availability of text prompts and technical quality considerations. In~\cite{Agin}, the proposed AGIN dataset was used to individually learn technical quality and rationality aspects, which were then effectively combined to develop an objective image naturalness metric. The resulting models, JOINT and JOINT+, achieved SROCCs of 0.8173 and 0.8351, respectively, with respect to the corresponding ground truth MOSs.


\section{Dataset Construction}

In this paper, we introduce a dataset for Realness Assessment for Image Synthesis and Evaluation (RAISE), which comprises subjective realness scores for 600 images. Unlike existing datasets such as~\cite{li2023agiqa, wang2023aigciqa2023, li2024aigiqa}, which provide perceptual quality scores, RAISE focuses specifically on perceptual realness—a distinct visual concept. While perceptual quality is influenced by technical factors like distortions or visual artifacts, perceptual realness is dictated by the plausibility of the depicted scene, logical object placement, and natural appearance of colors and textures. Although quality degradation may affect perceived realness, they play a smaller role compared to semantic and contextual coherence, as shown in \cite{Agin}. Indeed, camera captured images often contain distortions that lower technical quality without necessarily reducing perceived realness. To explore this relationship, RAISE also includes real images, allowing us to collect and compare realness ratings for both real and synthetic content. We argue that perceptual realness should be studied independently of technical quality, especially given that established IQA/VQA models for natural images can be readily applied to AIGIs to assess their technical quality when such AIGIs are realistic, i.e. they share similar distribution with real images. Thus, the purpose of the RAISE dataset is to facilitate the development and benchmarking of objective realness assessment models for AIGIs.

Furthermore, our subjective study was designed to be prompt agnostic, as AIGIs may not always have associated text prompts. Although~\cite{wang2023aigciqa2023} and~\cite{Agin} also provide naturalness ratings in addition to subjective quality ratings, \cite{wang2023aigciqa2023} included text prompts in the evaluation procedure, while~\cite{Agin} involved collecting discrete, categorical subjective opinions about naturalness, technical quality and rationality in contrast to the continuous scale of ratings adopted in our subjective study. 

Thus, the RAISE dataset is distinct from prior benchmarks such as AGIQA-1K \cite{zhang2023perceptual}, AIGIQA-3K~\cite{li2023agiqa}, AGIN~\cite{Agin}, and AIGIQA-20K~\cite{li2024aigiqa} in the following key aspects:

\begin{enumerate}
    \item A dedicated subjective evaluation framework focused on perceptual realness rather than technical image quality.
    \item A prompt-agnostic rating methodology, ensuring applicability to AIGIs irrespective of availability of associated text prompts.
    \item Inclusion of real photographs in the evaluation set to serve as implicit reference anchors for benchmarking the realness of AIGIs.
\end{enumerate}

\subsection{Content Collection}

\begin{figure}[h]
	\centering
	\includegraphics[width=0.85\linewidth]{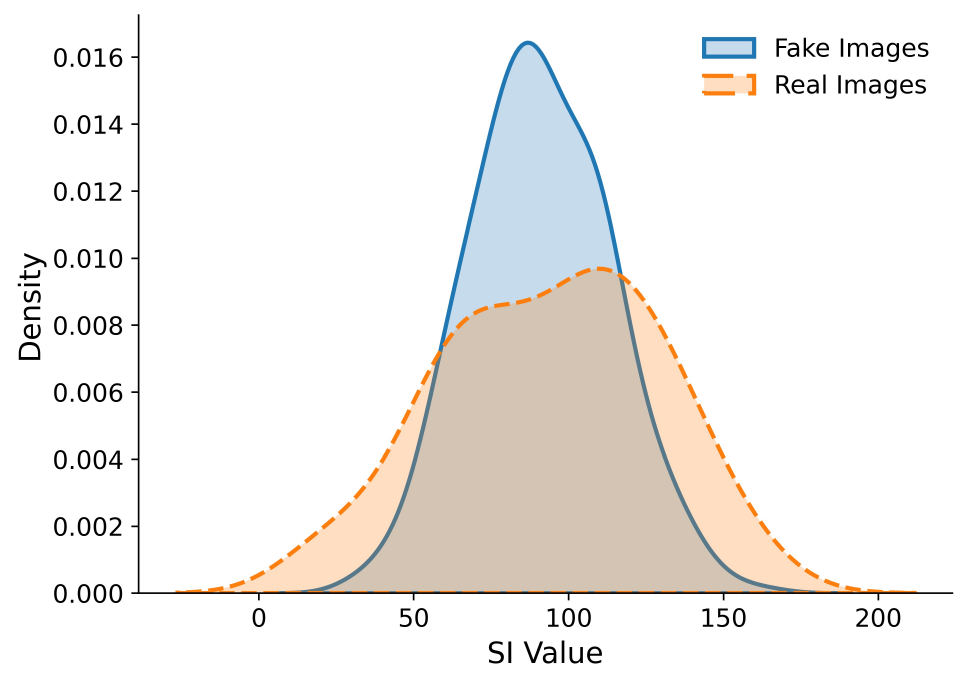}
	\caption{Distribution of SI values in RAISE visualized using kernel density estimate (KDE) plot.}
	\label{fig:SI-dist}
\end{figure}

The RAISE dataset consists of 480 AIGIs and 120 real images. The AIGIs present in our dataset were generated using Stable Diffusion v1.5 \cite{latentdiff} and were sourced from the Fake2M \cite{Fake2M} dataset. The real images were collected from KADID \cite{kadid}, KonIQ \cite{koniq} and CSIQ \cite{csiq} datasets. The resolution of all the images in our dataset is 512$\times$384 pixels.

Both the real and AI generated images were carefully curated to encompass a broad range of scenes. Specifically, as human faces are highly photographed and salient visual entities that are also particularly vulnerable to AI generated visual distortions, we ensured that the AIGI content of our dataset has a balanced representation of images depicting human faces. AIGIs were manually chosen to span a wide range of perceptual realness levels. The FID score calculated between the real and AIGI subsets of our dataset is 177.07. Both the real and the AIGI subsets of RAISE exhibit a wide range of scene complexities, as illustrated by the distribution of the spatial information (SI) \cite{itu910} values in Fig. \ref{fig:SI-dist}

\subsection{Design of Subjective Study} 

\begin{figure}[h]
	\centering
	\includegraphics[width=0.7\linewidth]{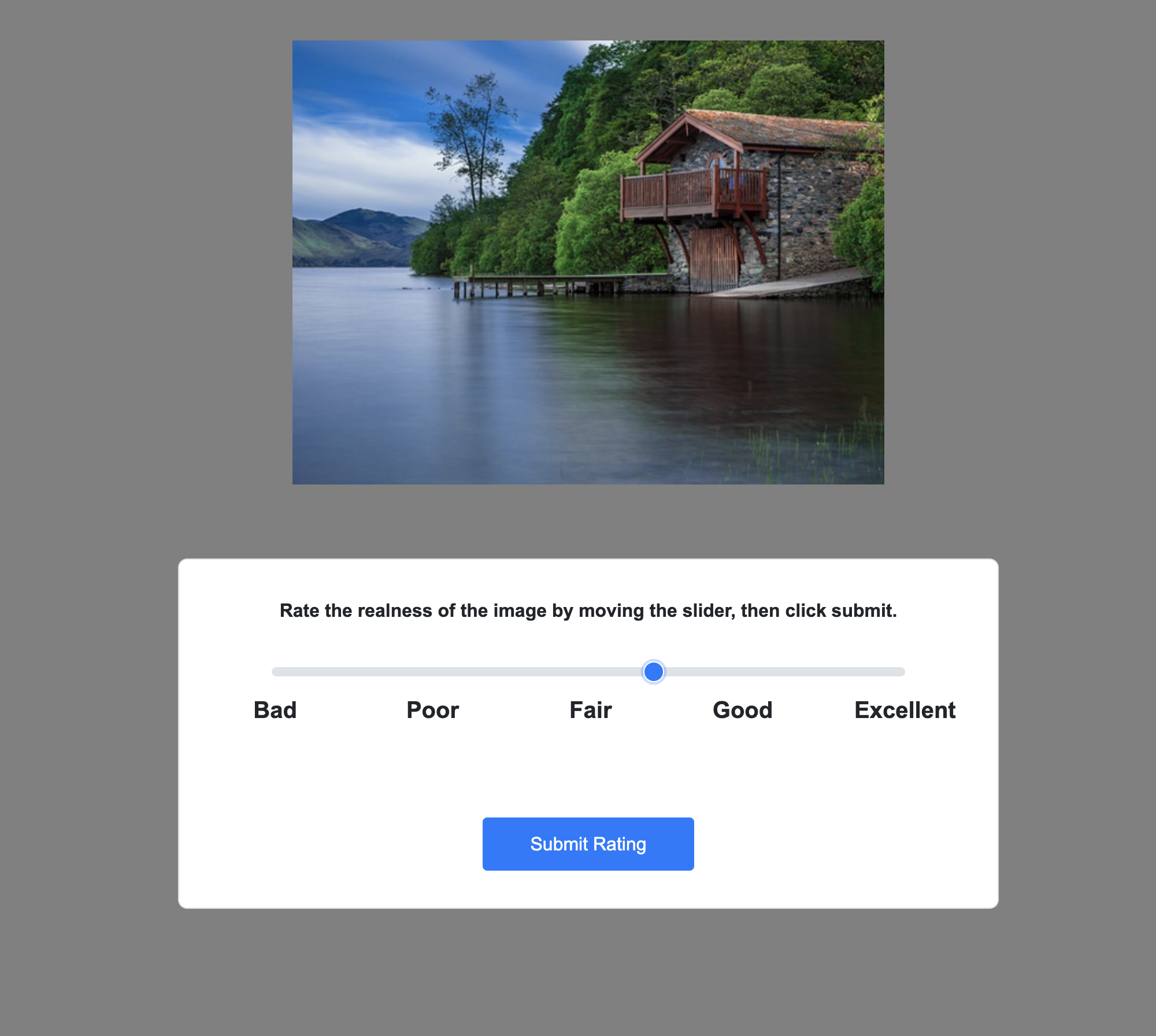}
	\caption{Rating screen shown to the participants of the human study.}
	\label{fig:rating-screen}
\end{figure}

We designed a psychovisual human study to collect reliable subjective ratings of the perceived realness following the ITU recommendation outlined in \cite{iturbt500}. The human study was conducted using a single stimulus testing methodology, where each participant rated the perceived realness of a single image--either real or AIGI--on a continuous scale ranging from 0-100, labeled with absolute category rating (ACR) labels to describe the realness levels. The rating interface is demonstrated in Fig. \ref{fig:rating-screen}. 

The subjective ratings were collected over two untimed  sessions, each lasting less than 30 minutes. At the beginning of the first session, each participant underwent a short training session to accustom them to the rating interface using 5 sample images (not included in the RAISE dataset) selected to broadly correspond to the 5 ACR levels of realness (`Bad', `Poor', `Fair', `Good' and `Excellent'). 150 images were presented to each participant in a single session using a randomized round-robin schedule. 

A total of 47 subjects participated in the study, comprising 41 men and 6 women, aged between 20 and 31 years (mean age: 23.5 years). Additional participant details that were obtained using standard tests for visual acuity and color vision as well as through a pre-study questionnaire about prior exposure to AI generated imagery are reported in Table \ref{tab:participant_stats}. 

\begin{table}[h!]
\centering
\caption{Details of Participants of the Human Study.}
\begin{tabular}{|l|c|}
\hline
\textbf{Characteristic} & \textbf{Number} \\ \hline
Color vision & 45 \\ \hline
Normal/Corrected to Normal Visual Acuity & 44  \\ \hline 
Prior exposure to AIGC  &  28 \\ \hline
\end{tabular}
\label{tab:participant_stats}
\end{table}


The human study described in this section resulted in 14,100 total ratings, with each image's realness assessed by at least 23 distinct participants.

\section{Statistical Analysis of Subjective Ratings}

We conducted a kurtosis-based post-screening procedure as recommended in \cite{iturbt500} to detect and exclude unreliable raters. Based on this criterion, ratings from three subjects were identified as outliers and subsequently discarded. The MOS for each image in the RAISE dataset was computed by averaging the remaining participants' ratings. The inter-subject consistency, measured using SROCC, was found to be 0.7327, indicating substantial agreement among participants in their realness assessments.

\begin{figure}[h]
    \centering
    \includegraphics[width=0.88\linewidth]{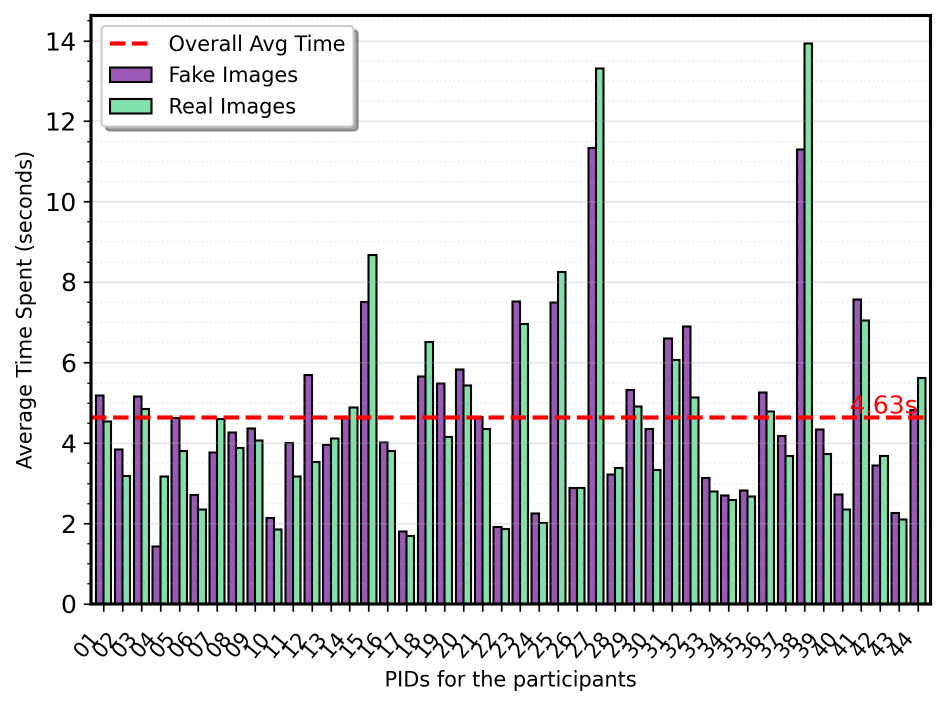}
    \caption{Average amount of time taken by subjects to rate images.}
    \label{fig:time-taken}
\end{figure}

The average time taken by each participant to rate the images is shown in Fig.~\ref{fig:time-taken}. From this figure, we observe that the rating times for real and AI-generated images were quite similar across subjects. On average, participants took 4.54 seconds to rate real images and 4.64 seconds for AI-generated images.

\begin{figure}[h]
	\centering
	\includegraphics[width=0.72\linewidth]{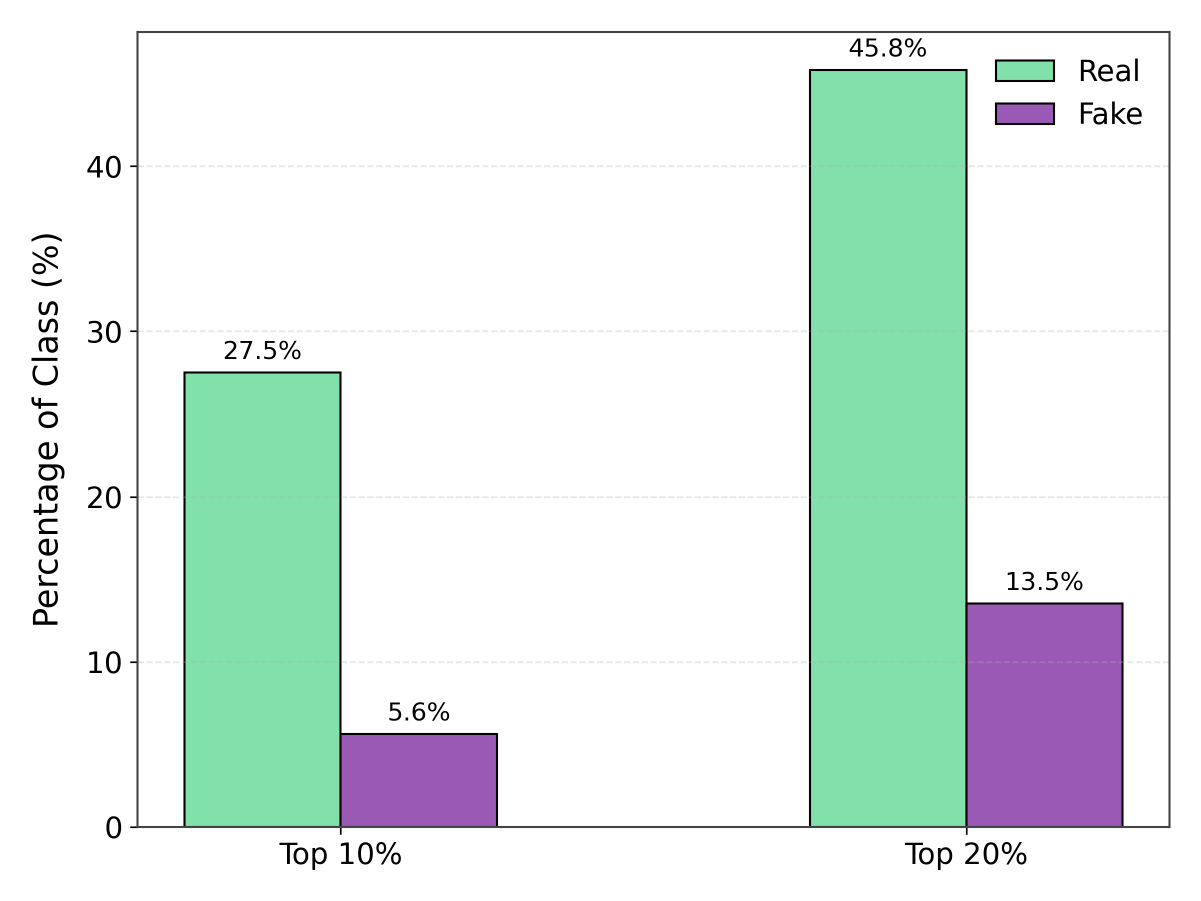}
	\caption{Bar graphs showing the proportion of real and fake images in the top 10 and 20 percentiles of MOS ratings.}
	\label{fig:percentiles}
\end{figure}

To further analyze the distribution of realness scores, we examine the proportion of real and AI-generated images (AIGIs) appearing in the top 10 and top 20 percentiles of MOS realness scores, as shown in Fig.~\ref{fig:percentiles}. Specifically, 27.5\% and 45.8\% of the real images fall within the top 10 and top 20 percentiles of scores, respectively, compared to only 5.6\% and 13.5\% of AIGIs. These statistics align with the expectation that real images are generally perceived as more realistic than AIGIs. However, since the AIGIs in our dataset were selected to span a broad range of realness levels, a small fraction achieved sufficiently high ratings to appear in the top 10 percentile.

\begin{figure}[h]
    \centering
    \includegraphics[width=0.88\linewidth]{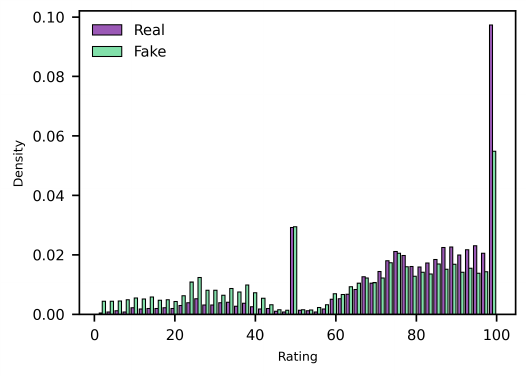}
    \caption{Distribution of MOS ratings on the real and AIGI subsets.}
    \label{fig:rating-dist}
\end{figure}

The distribution of MOS realness scores is shown in Fig. \ref{fig:rating-dist}, with bar heights for real and AIGI 
subsets normalized by the number of images in the respective subset. Real images achieved an average MOS of 76.56, while AIGIs averaged 60.98. Fig. \ref{fig:rating-dist} reveals that both real and synthetic images span the full range of the rating scale. Notably, this indicates that even real images can be perceived as synthetic by human raters, compounding the challenge of achieving reliable separation between the two subsets.

\begin{figure}[h]
	\centering
	\includegraphics[width=0.92\linewidth]{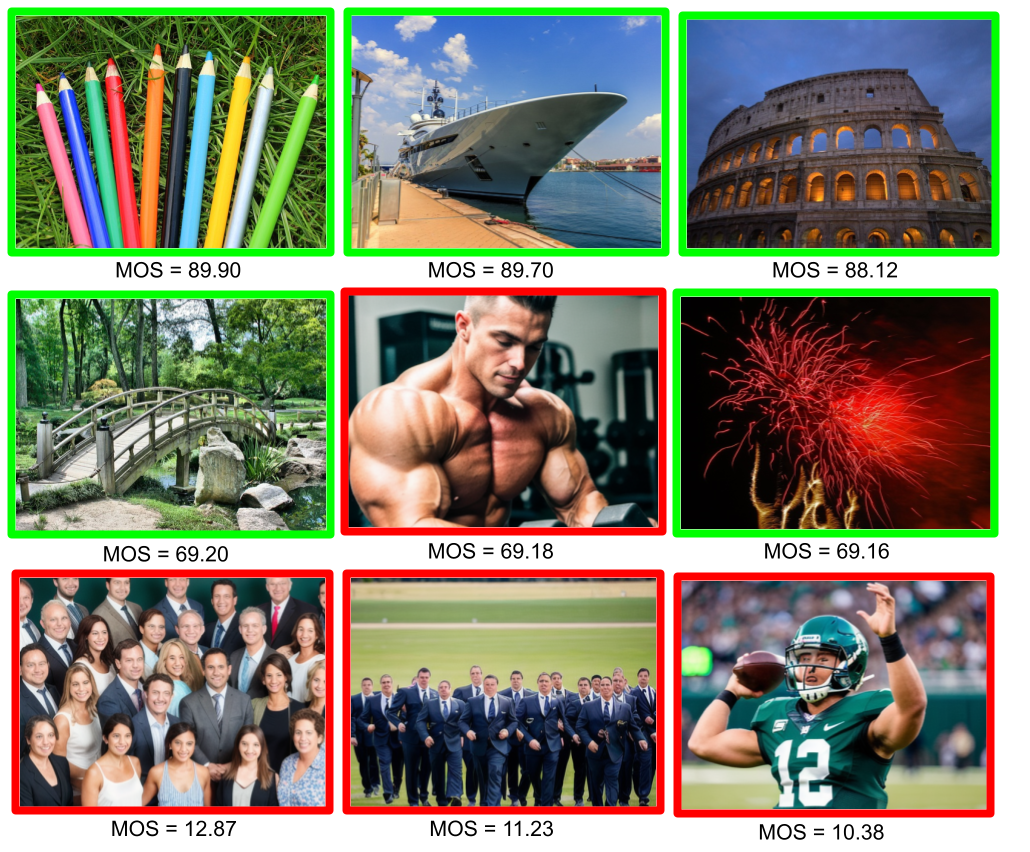}
	\caption{Images from the RAISE dataset corresponding to the highest, median and lowest MOS ratings. The images outlined in green are real images while the ones outlined in red are AI generated.}
	\label{fig:raise_images}
\end{figure}

Fig. \ref{fig:raise_images} shows sample images from the RAISE dataset with the three images closest to the highest, median, and lowest MOS realness scores shown in the top, middle, and bottom rows, respectively. The lowest-rated images, all AIGIs, exhibit decisively unrealistic visual attributes, leading to low realness scores. In contrast, the highest-rated images were all real and consistently judged as realistic. The images rated near the median MOS score include both real and synthetic content, with greater variability in subjective ratings, reflecting the ambiguity in their perceived realness.

\section{Modeling Objective Realness}

In this section, we describe a number of approaches that we employed to objectively model the subjective ratings in our newly introduced RAISE dataset. To train and evaluate the models elucidated in this section, we partitioned the RAISE dataset into training and test partitions having 510 and 90 images, respectively, ensuring that the distribution of MOS ratings were consistent across the two partitions. 


\subsection{Feature Design}
\label{subsec:ML_models}

We first considered several low and mid-level image features that could influence human perception of realness. These features are organized into five categories, described as follows. 
\begin{enumerate}
    \item \textit{Color Features}: In natural images, color components tend to be highly correlated, whereas generative AI models often produce implausible color relationships. AIGIs frequently exhibit oversaturated or muted color palettes and unnatural exposure characteristics. We model these effects using both HSV and Lab color spaces, which decouple chromatic components from luminance information \cite{amin2024exploring}.
    \item \textit{Edge Features}: AIGIs often contain geometrical inconsistencies that manifest as irregular or unnatural edge orientations \cite{chintha2020leveraging}. We extract edge features using the Sobel operator to capture these anomalies 
   \item \textit{Contrast Features}: Natural images typically display structured contrast patterns, while AIGIs may deviate from these norms. We capture contrast variations using contrast energy derived from Gaussian bandpass filter responses and from Gaussian gradient magnitudes.
    \item \textit{Texture and Structure Features}: AIGI are often characterized by unnatural textures such as unusual smoothness, repetitive patterns, and inconsistent spatial structures \cite{xu2021deepfake}. We derive texture descriptors from gray level co-occurrence matrix (GLCM) \cite{glcm} and structure descriptors from histogram of oriented gradients (HoG) \cite{Hog} to account for such effects \cite{kharbat2019image}. 
    \item \textit{Frequency Features}: generative AI often introduces atypical frequency signatures in images \cite{frank2020leveraging}, an effect which capture by using features derived from a 2D wavelet decomposition of the images using Daubechies wavelets. 

\end{enumerate}

We trained various machine learning models using 55 handcrafted features across all the categories described above, computed from images in the training partition of the RAISE dataset. Among the models evaluated, a decision tree regressor yielded the best performance. Table~\ref{tab:feature_based_models} reports the average performance of this regressor on the test partition of the RAISE dataset. The model achieved a SROCC of 0.5033 and a Pearson's linear correlation coefficient (PLCC) of 0.4339, as shown in the last row of the table. These results demonstrate that even a compact set of relatively simple, low and mid-level features can capture perceptual cues influencing human judgments of image realness to a reasonable extent.

We also performed an ablation study by training the decision tree regressor while excluding one feature category at a time. Rows 1–5 in Table~\ref{tab:feature_based_models} report the SROCC and PLCC values obtained by omitting the feature category listed in the first column of each row. The consistent drop in performance in all cases evinces that each feature group contributes significantly to the model’s ability to predict perceptual realness scores.

\begin{table}[h]
\centering
\caption{Performance of low and mid level image features on RAISE.}
\resizebox{0.95\linewidth}{!}{  
\begin{tabular}{|c|c|cc|}
\hline
\textbf{Category} & \makecell{\textbf{\# of} \\ \textbf{Features}} & \multicolumn{2}{c|}{\makecell{\textbf{Performance} \\ \textbf{(excluding col. 1  for rows 1--5)}}} \\
           &            & \textbf{SROCC} & \textbf{PLCC} \\
\hline
Color & 8 & \textbf{0.5106} & 0.3922 \\
Edge & 10 & 0.3417 & 0.3167 \\
Contrast & 3 & 0.4887 & \textbf{0.4833} \\
Texture \& Structure & 22 & 0.3326 & 0.3901 \\
Frequency & 12 & 0.4622 & 0.4260 \\
All & 55 & 0.5033 & 0.4339  \\
\hline
\end{tabular}
}
\label{tab:feature_based_models}
\end{table}

\subsection{CNN-based Feature Learning}
\label{subsec:CNN_model}

To obtain better predictions of subjective realness scores, we trained a convolutional neural network (CNN) in a supervised manner using the images and the corresponding subjective MOS ratings of realness from the RAISE dataset. Considering that the RAISE dataset has just 600 images, we used a lightweight CNN architecture that consisted of 5 convolutional layers and 3 linear layers, with a total of 434,817 trainable parameters. To avoid overfitting, we applied regularization techniques such as dropout \cite{dropout} for linear layers and batch normalization \cite{batchnorm} for convolutional layers, along with data augmentation (randomized flips, rotations, cropping followed by resizing and color jitter).

The trained CNN model achieved an SROCC of 0.6028 and a PLCC of 0.6490 on the test set. Thus, it surpassed the performance of the handcrafted feature-based model in Section \ref{subsec:ML_models} indicating that, despite the limited sample size, the CNN successfully learned representative features that captured visual naturalness and inconsistencies in the images more effectively.

\subsection{Transfer Learning}
\label{subsec:transfer_learning}

Transfer learning with foundation vision models is particularly effective when annotated data is limited, as in our case. For the purpose of transfer learning, we selected the ResNet-18 \cite{resnet} backbone pretrained on ImageNet \cite{imagenet}, which is well-suited for feature extraction across diverse vision tasks. We used the 512-dimensional feature map obtained by performing global average pooling of the deep features produced by the last residual block of ResNet-18, as the input to a neural network (NN) with 4 fully connected layers. This NN regressor consisted of 230,145 trainable parameters. We used dropout and data augmentation for regularization as described in \ref{subsec:CNN_model} and the parameters of the ResNet-18 backbone remained unchanged during training.

By using ResNet-18 as a pretrained feature extractor to train a NN regressor, we obtained an SROCC of 0.6684 and a PLCC of 0.7395 on the test set. Thus, this model trained with transfer learning conclusively outperformed the CNN  trained from scratch as elaborated in Section \ref{subsec:CNN_model}, despite having 47.1\% fewer parameters. These results thus attests to the efficacy of foundation vision models such as ResNet-18 trained on large datasets as pretrained feature extractor for quantifying perceptual realness. 

\subsection{JOINT}

Since the JOINT model \cite{Agin} was trained on subjective opinions of rationality and technical quality to predict overall AIGI naturalness, it is pertinent to evaluate its performance on the RAISE dataset introduced in this paper. We thus conducted inference on the RAISE test set with the trained JOINT model with official weights \footnote{Official JOINT implementation: \url{https://github.com/zijianchen98/AGIN}}. Since the JOINT model predicts naturalness scores in the range of 1-5, we compared the model's predictions with the subjective MOS ratings by scaling the MOS ratings to the same range. 

The performance of the JOINT model on RAISE is reported in Table \ref{tab:joint_performance}. The first two rows show the SROCC and PLCC for the scores predicted by the technical quality and rationality branches, respectively, using the JOINT parameters trained on the AGIN dataset. The third row presents the results obtained by a weighted average of both branches, with a weight of 0.769 for the rationality branch and 0.145 for the technical quality branch as suggested in \cite{Agin}. The overall model achieved an SROCC of 0.2998 and a PLCC of 0.2776 on our test set. Although our MOS ratings reflect perceived realness, the technical quality branch correlated better with the data than the rationality branch. Despite being trained on the significantly larger AGIN dataset, JOINT failed to generalize effectively to RAISE, performing worse than a simple decision tree regressor trained on handcrafted features. This result highlights the limitations of existing models in capturing subjective realness, underscoring the need for more robust approaches to realness prediction for AIGIs.

\begin{table}[h]
\centering
\caption{Performance of JOINT on the RAISE dataset.}
\resizebox{0.95\linewidth}{!}{  
\begin{tabular}{|l|c|c|}
\hline
\textbf{Model} & \textbf{SROCC} & \textbf{PLCC} \\
\hline
Pretrained Technical Quality       & 0.3298 & 0.3518 \\
Pretrained Rationality             & 0.2319 & 0.2156 \\
Overall Pretrained                 & 0.2998 & 0.2776 \\
Transfer learning (without fine tuning) & 0.6785 & 0.7027 \\
Transfer learning (with fine tuning)  & \textbf{0.6798} & \textbf{0.7421} \\
\hline
\end{tabular}}
\label{tab:joint_performance}
\end{table}

We next applied transfer learning using the pretrained JOINT model parameters. Since our dataset contains subjective realness annotations, we utilized the rationality branch of JOINT which incorporates a fine-tuned ResNet-50 backbone pretrained on the AVA dataset \cite{ava}. We used the learned weights of JOINT to perform feature extraction with this backbone network, and trained a fully connected NN regressor having 32,582,501 parameters with dropout and data augmentation regularization. This transfer learning approach yielded an SROCC of 0.6785 and a PLCC of 0.7027 on the test set, as reported in the 4th row of Table \ref{tab:joint_performance} While the SROCC improved, the PLCC declined relative to transfer learning with ResNet-18 initialized with the default set of parameters trained on ImageNet \cite{imagenet}

\begin{figure}[htb]
	\centering
	\includegraphics[width=0.75\linewidth]{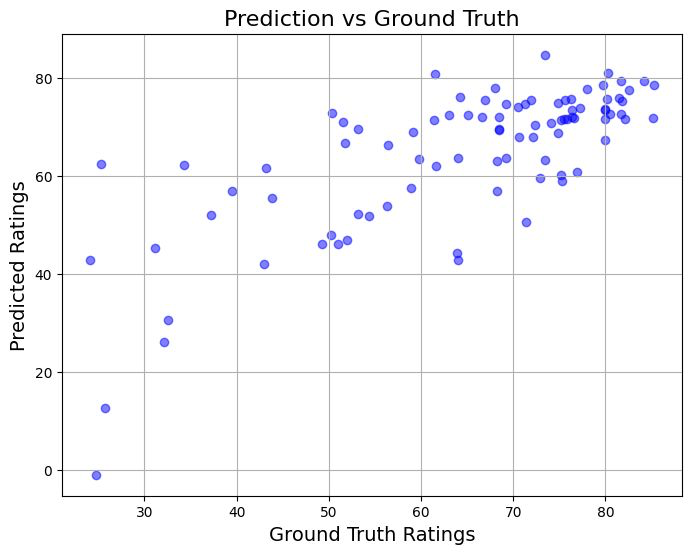}
	\caption{Scatter plot of the predictions obtained by fine-tuning the last residual block of JOINT rationality branch against ground truth MOS ratings.}
	\label{fig:scatter_plot}
\end{figure}

We further fine-tuned the parameters of all the layers of the last residual blocks of the rationality branch of JOINT while training the fully connected NN regressor from scratch, resulting in a total of 47,547,237 parameters, train with dropout and data augmentation regularization. The resulting performance is reported in the last row of Table \ref{tab:joint_performance}. The SROCC improved to 0.6798, while the PLCC improved to 0.7421. A scatter plot of the predicted versus ground-truth MOS scores obtained with this model is shown in Fig. \ref{fig:scatter_plot}. These results demonstrate that although the pretrained JOINT model performed poorly on the RAISE dataset, the features learned by its rationality branch significantly improved prediction performance when adapted through partial fine-tuning.

The performance of all models discussed in this section is summarized in Table \ref{tab:summary}. These models establish baseline performances on our dataset, providing a foundation for developing more advanced methods leveraging feature design, representation learning, or transfer learning.

\begin{table}[h]
\centering
\caption{Model Performance Summary.}
\resizebox{0.99\linewidth}{!}{
\begin{tabular}{|c|c|cc|}
\hline
\textbf{Model} & \makecell{\textbf{\# of} \\ \textbf{Parameters} \\ \textbf{Trained}} & \multicolumn{2}{c|}{\textbf{Performance}} \\
& & \textbf{SROCC} & \textbf{PLCC} \\
\hline
Decision Tree & 0 (Non-parametric) & 0.5106 & 0.3922 \\ \hline
CNN & 434,817 & 0.6028 & 0.6490 \\ \hline
NN on ResNet-18 features & 230,145 & 0.6684 & 0.7395 \\ \hline
JOINT & 0 (Pretrained) & 0.2998 & 0.2776 \\ \hline
\makecell{NN on fine-tuned JOINT \\ rationality features} & 47,547,237 & \textbf{0.6798} & \textbf{0.7421} \\
\hline
\end{tabular}
}
\label{tab:summary}
\end{table}


\section{Conclusion}
In this paper, we introduced the RAISE dataset, consisting of AI-generated and real images along with subjective realness ratings measured as MOS values. We analyzed the dataset's statistics and established baseline performances of several models using traditional feature-based and deep learning approaches. As future work we aim to develop more advanced models for realness prediction that incorporate perceptual mechanisms such as visual attention. Additionally, we will focus on developing learned models for the joint localization of visual inconsistencies and realness prediction. Ultimately, we plan to integrate the objective realness assessment and localization results into the learning framework of generative AI models to reduce their visual inconsistencies and unrealistic artifacts. 

\balance
\bibliographystyle{IEEEtran}
\bibliography{bibliography.bib}

\end{document}